\pdfoutput=1

\documentclass[11pt]{article}

\usepackage{ACL2023}

\usepackage{times}
\usepackage{latexsym}

\usepackage[T1]{fontenc}

\usepackage[utf8]{inputenc}

\usepackage{microtype}

\usepackage{inconsolata}

\usepackage{todonotes}
\usepackage{cleveref}
\usepackage[british]{babel}  
\usepackage[nolist,nohyperlinks]{acronym}
\usepackage{multirow}  
\usepackage{array}
\usepackage{booktabs}
\usepackage{longtable}

\title{An Open Dataset and Model for Language Identification}

\author{Laurie Burchell \and Alexandra Birch \and Nikolay Bogoychev \and Kenneth Heafield \\
    Institute for Language, Cognition, and Computation \\
    School of Informatics, University of Edinburgh \\
    10 Crichton Street, Edinburgh, EH8 9AB, UK \\
    \texttt{\{laurie.burchell,a.birch,n.bogoych,kenneth.heafield\}@ed.ac.uk} 
}

\begin{acronym}
    \acro{fpr}[FPR]{false positive rate}
    \acro{lid}[LID]{language identification}
    \acro{nlp}[NLP]{natural language processing}
    \acro{msa}[MSA]{Modern Standard Arabic}
    \acro{udhr}[UDHR]{Universal Declaration of Human Rights}
\end{acronym}

\begin{document}

\maketitle

\begin{abstract}

\Acf{lid} is a fundamental step in many \acl{nlp} pipelines. However, current \ac{lid} systems are far from perfect, particularly on lower-resource languages. We present a \ac{lid} model which achieves a macro-average F1 score of 0.93 and a false positive rate of 0.033 across 201 languages, outperforming previous work. We achieve this by training on a curated dataset of monolingual data, the reliability of which we ensure by auditing a sample from each source and each language manually. We make both the model and the dataset available to the research community. Finally, we carry out detailed analysis into our model's performance, both in comparison to existing open models and by language class.

\end{abstract}

\section{Introduction}
\Acf{lid} is a foundational step in many \ac{nlp} pipelines. It is used not only to select data in the relevant language but also to exclude `noise'. For this reason, effective \ac{lid} systems are key for building useful and representative \ac{nlp} applications.

Despite their importance, recent work has found that existing \ac{lid} algorithms perform poorly in practice compared to test performance \citep{caswell-etal-2020-language}. The problem is particularly acute for low-resource languages: \citet{kreutzer-etal-2022-quality} found a positive Spearman rank correlation between quality of data and size of language for all of the \ac{lid}-filtered multilingual datasets they studied. In addition, for a significant fraction of the language corpora they studied, less than half of the sentences were in the correct language. They point out that such low-quality data not only leads to poor performance in downstream tasks, but that it also contributes to `representation washing', where the community is given a false view of the actual progress of low-resource \ac{nlp}.

For applications such as corpus filtering, \ac{lid} systems need to be fast, reliable, and cover as many languages as possible. There are several open \ac{lid} models offering quick classification and high language coverage, such as CLD3 or the work of  \citet{costa2022no}. However, to the best of our knowledge, none of the commonly-used scalable \ac{lid} systems make their training data public. This paper addresses this gap through the following contributions:

\begin{itemize}
    \item We provide a curated and open dataset covering 201 languages. We audit a sample from each source and each language making up this dataset manually to ensure quality.
    \item We train a \ac{lid} model on this dataset which outperforms previous open models. We make this model publicly available.\footnote{\url{github.com/laurieburchell/open-lid-dataset}}
    \item We analyse our model and use our findings to highlight open problems in \ac{lid} research.
\end{itemize}

\section{Background}
There is a long history of research into \ac{lid} using a plethora of methods \citep{jauhiainen2019automatic}. For high-coverage \ac{lid}, \citet{dunn2020mapping} presents a model covering 464 languages, whilst \citet{brown-2014-non} includes as many as 1366 language varieties. Unlike our work, the training data in both cases has not been manually checked for quality. Recent work by \citet{adebara2022afrolid} presents a LID system covering 517 African languages and varieties where the training data has been curated manually. However, as far as we are aware this data is not easily available.

\citet{costa2022no} released a substantial piece of research aiming to improve machine translation coverage for over 200 languages. As part of this, they provided several professionally-translated datasets for use as test and development sets. For this reason, we use their system as our benchmark. However, whilst they did release scripts to recreate their parallel data,\footnote{\url{github.com/facebookresearch/fairseq/tree/nllb}} they did not provide---or even document---the monolingual data used to train their \ac{lid} system, saying only that they use ``publicly available datasets'' supplemented with their own dataset NLLB-Seed. By providing an open dataset, we aim to facilitate futher research.

\section{Dataset}

\subsection{Data sources}
We wanted to be as confident as possible that our dataset had reliable language labels, so as to avoid the problems noted in existing corpora \citep{kreutzer-etal-2022-quality}. We therefore avoided web-crawled datasets and instead chose sources where we felt the collection methodology made it very likely that the language labels were correct.

The majority of our source datasets were derived from news sites, Wikipedia, or religious text, though some come from other domains (e.g. transcribed conversations, literature, or social media). A drawback of this approach is that most of the text is in a formal style. Further work could collect data from a wider range of domains whilst maintaining trust in the labels. We checked that each dataset was either under an open license for research purposes or described as free to use. A full list of sources is given in \Cref{sec:sources}, and further information including licenses is available in the code repository accompanying this paper.

\subsubsection{Language selection}
\label{sec:coverage}
Our initial aim was to cover the same languages present in the FLORES-200 Evaluation Benchmark\footnote{\url{github.com/facebookresearch/flores/blob/main/flores200}} so that we could use this dataset for evaluation and compare our results directly with \citet{costa2022no}. However, during the curation process, we decided to exclude three languages. 

Firstly, though Akan and Twi are both included as separate languages in FLORES-200, Akan is actually a macrolanguage covering a language continuum which includes Twi. Given the other languages in FLORES-200 are individual languages, we decided to exclude Akan.

Secondly, FLORES-200 includes \ac{msa} written in Latin script. It is true that Arabic dialects are often written in Latin characters in informal situations (e.g. social media). However, \ac{msa} is a form of standardised Arabic which is not usually used in informal situations. Since we could not any find naturally-occurring training data, we excluded \ac{msa} from the dataset.

Finally, we excluded Minangkabau in Arabic script because it is now rarely written this way, making it difficult to find useful training data.\footnote{\url{omniglot.com/writing/minangkabau.htm}, \url{ethnologue.com/language/min}}

\subsection{Manual audit process}
The first step in our manual audit was to check and standardise language labels, as these are often inconsistent or idiosyncratic \citep{kreutzer-etal-2022-quality}. We chose to copy the language codes in \citet{costa2022no}, and reassign macrolanguage or ambiguous language codes in the data sources we found to the dominant individual language. Whilst this resulted in more useful data for some languages, for other languages we had to be more conservative. For example, we originally reassigned text labelled as the macrolanguage Malay (\textit{msa\_Latn}) to Standard Malay, but this led to a large drop in performance as the former covers a very diverse set of languages.

Two of the authors then carried out a manual audit of a random sample of all data sources and languages:\footnote{Specifically, we used the following command on each file to select 500 lines to audit: \texttt{`shuf <file> | head -n 500 | less}} one a native Bulgarian speaker (able to read Cyrillic and Latin scripts and Chinese characters), and the other a native English speaker (able to read Latin, Arabic and Hebrew scripts). For languages we knew, we checked the language was what we expected. For unfamiliar languages in a script we could read, we compared the sample to the \ac{udhr} or failing that, to a sample of text on Wikipedia. We compared features of the text which are common in previous \ac{lid} algorithms and could be identified easily by humans: similar diacritics, word lengths, common words, loan words matching the right cultural background, similar suffixes and prefixes, and vowel/consonant patterns \citep[Section 5]{jauhiainen2019automatic}. For scripts we could not read, we checked that all lines of the sample matched the script in the \ac{udhr}. 

\subsection{Preprocessing}
We kept preprocessing minimal so that the process was as language agnostic as possible. We used the scripts provided with Moses \citep{koehn-etal-2007-moses} to remove non-printing characters and detokenise the data where necessary. We then filtered the data so that each line contained at least one character in the expected script (as defined by Perl) to allow for borrowings. Finally, we followed \citet{arivazhagan2019massively} and \citet{costa2022no} and sampled proportionally to $ p_l^{0.3} $, where $ p_l $ is the fraction of lines in the dataset which are in language $ l $. This aims to ameliorate class skew issues.

\subsection{Dataset description}
The final dataset contains 121 million lines of data in 201 language classes. Before sampling, the mean number of lines per language is 602,812. The smallest class contains 532 lines of data (South Azerbaijani) and the largest contains 7.5 million lines of data (English). There is a full breakdown of lines of training data by language in \Cref{sec:by_lang_perf}.

\section{Model and hardware}
We used our open dataset to train a \textit{fasttext} \ac{lid} model using the command-line tool \citep{joulin-etal-2017-bag}. It embeds character-level n-grams from the input text, and then uses these as input to a multiclass linear classifier. We used the same hyperparameters as \citet{costa2022no} (NLLB), which we list in \Cref{sec:hyperparams}. We trained our model on one Ice Lake node of the CSD3 HPC service. Each node has 76 CPUs and 256GiB of RAM. Our model takes c. 1hr 45mins to train and contains 60.5 million parameters. Inference over the 206,448 lines of the test set takes 22.4 secs (9216.4 lines/sec).

\begin{table*}[htbp]
    \small
    \centering
    \begin{tabular}{cccccccc}
        & & \multicolumn{2}{c}{FLORES-200$^*$} & \multicolumn{2}{c}{FLORES200$^*\cap$ NLLB} & \multicolumn{2}{c}{FLORES-200$^*\cap$ CLD3} \\
        & & \multicolumn{2}{c}{201 languages} & \multicolumn{2}{c}{193 languages} & \multicolumn{2}{c}{95 languages} \\
        \cmidrule(lr){3-4} \cmidrule(lr){5-6} \cmidrule(lr){7-8}
        System & Supported languages. & F1 $ \uparrow $ & FPR $ \downarrow $ & F1 $ \uparrow $ & FPR $ \downarrow $ & F1 $ \uparrow $ & FPR $ \downarrow $ \\
        \midrule
        CLD3 & 107  & - & - & - & -  & 0.968 & 0.030 \\
        NLLB  & 218  & - & -  & 0.950 & 0.023 & 0.985 & 0.019 \\
        Our model & 201 & \textbf{0.927} & \textbf{0.033} & \textbf{0.959} & \textbf{0.020} & \textbf{0.989} & \textbf{0.011} \\
         
    \end{tabular}
    \caption{A comparison of open-source \ac{lid} systems.  \textit{Supported languages} gives the number of languages the classifier claims to support. Each column gives the classifier's performance on a test set containing the intersection of languages each classifier claims to support. We report macro-averages of F1 scores and \acp{fpr}.}
    \label{tab:results}
\end{table*}

\section{Evaluation}

\subsection{Test sets}
We use the FLORES-200 benchmark provided by \citet{costa2022no} for evaluation. It consists of 842 distinct web articles sourced from English-language Wikimedia projects, with each sentence professionally translated into 204 languages. The target side is human-verified as in the right language, making it suitable for use as a \ac{lid} evaluation set. For each language, 997 sentences are available for development and 1012 for dev-test (our test set).\footnote{992 sentences are withheld by \citet{costa2022no} as a hidden test set.} We remove the three languages discussed in \Cref{sec:coverage} from FLORES-200, leaving 201 languages in the test set: FLORES-200$^*$.

\subsection{Other \ac{lid} systems}

We compare our model's performance to two other open-source \ac{lid} systems: \texttt{nllb218e} (NLLB)\footnote{\url{tinyurl.com/nllblid218e}} and \texttt{pycld3 0.22} (CLD3).\footnote{\url{pypi.org/project/pycld3}} We discuss how we ensured a fair comparison below.

\textbf{NLLB} is a \textit{fasttext} model. We were surprised to discover that whilst it does cover 218 languages, it only includes 193 of the 201 languages in FLORES-200$^*$. This is despite the fact that the NLLB \ac{lid} model and the original FLORES-200 evaluation set were created as part of the same work \citep{costa2022no}. Referring to the analysis in the original paper, the authors note that ``Arabic languoids and Akan/Twi have been merged after linguistic analysis'' \citep[Table 5, p. 32]{costa2022no}. We discuss the reason to merge Akan and Twi in \Cref{sec:coverage}, but we judge Arabic dialects to be close but distinct languages. Our model performs poorly on Arabic dialects with the highest F1 score only 0.4894 (Moroccan Arabic). This is likely due to the general difficulty of distinguishing close languages combined with particularly sparse training data. We assume these poor results led to Arabic dialects (save \ac{msa}) being excluded from the NLLB \ac{lid} classifier. We remove eight Arabic dialects from the test set when comparing our model and NLLB, leaving 193 languages.

\textbf{CLD3} is an n-gram based neural network model for \ac{lid}. It uses different language codes to the other two models, so we normalise all predictions to  BCP-47 macrolanguage codes to allow fair comparison. We test on the 95 languages that all models have in common after normalisation.

\section{Results}
Our results are given in \Cref{tab:results}. We evaluate all models using F1 scores and \acf{fpr}. We report macro-averages to avoid down-weighting low-resource languages \citep{kreutzer-etal-2022-quality}. Following \citet{caswell-etal-2020-language}, we report \ac{fpr} to give a better indication of real-world performance when there is significant class skew.

We achieve an F1 score of 0.927 and a \ac{fpr} of 0.033 on FLORES-200$^*$. We also outperform both NLLB and CLD3 on the mutual subsets of FLORES-200$^*$. Since NLLB and our model share the same architecture and the same parameters, we attribute our success to our training data selection and manual audit process. 

Notably, our F1 score jumps to 0.959 and \ac{fpr} falls to 0.020 when we exclude the eight Arabic dialects from the test set to compare with NLLB. The 95 languages covered by CLD3, NLLB, and our model are mostly high resource, and so it is unsurprising that we achieve the highest F1 score (0.989) and lowest \ac{fpr} (0.011) on this subset.

We notice that the Pearson correlation between the number of lines of training data and F1 score for each language is only 0.0242. This is not unexpected: some of the least resourced languages achieve perfect scores on the test set due to high domain overlap, whereas the higher-resourced languages might get lower scores on the test set but have better robustness across domains. Full results by language are available in \Cref{sec:by_lang_perf}.

\subsection{Performance by language category} Using the taxonomy and list of languages in \citet{joshi-etal-2020-state}, we label each of the languages in our dataset according to its level of data availability (0 = least resourced, 5 = best resourced). We leave out 5 languages missing from the taxonomy, plus the 8 Arabic dialects not covered by NLLB. \Cref{tab:categories} compares the mean F1 score and \ac{fpr} of our model and for that of \citet{costa2022no} (NLLB). Our model has a higher or equal F1 score in every category and a lower or equal \ac{fpr} in every category but one, showing our model's improved performance across languages with different amounts of available data.

We note that class zero (the least-resourced languages) shows the smallest change in performance. We speculate that this is an artifact of the curation of our training dataset. For the best-resourced languages with more sources to choose from, it is likely that there is a significant difference between our training data and that used to train the model in \citet{costa2022no}. However, for the least-resourced languages, the sheer lack of resources means that overlap between our data and that used by \citet{costa2022no} is more likely. We suspect this is the reason we see little difference in performance for class zero in \Cref{tab:categories}. Unfortunately, without access to the training data used to train NLLB, we cannot verify this assumption.

\begin{table}[htbp]
    \small
    \centering
    \begin{tabular}{cccccc}
    & & \multicolumn{2}{c}{F1 $\uparrow$} & \multicolumn{2}{c}{FPR $\downarrow$} \\
    \cmidrule(lr){3-4} \cmidrule(lr){5-6}
    Class & Count & Ours & NLLB & Ours & NLLB \\  
    \midrule
    0 & 28 & \textbf{0.900} & 0.897 & 0.014 & \textbf{0.013} \\   
    1 & 94 & \textbf{0.981} & 0.968 & \textbf{0.013} & \textbf{0.013} \\
    2 & 16 & \textbf{0.990} & 0.963 & \textbf{0.009} & 0.043 \\
    3 & 25 & \textbf{0.983} & 0.974 & \textbf{0.007} & 0.013 \\
    4  & 18 & \textbf{0.951} & \textbf{0.951} & \textbf{0.051} & 0.055 \\
    5  & 7 & \textbf{0.897} & 0.855 & \textbf{0.163} & 0.620 \\
    \end{tabular}
    \caption{For each language class in the taxonomy of \citet{joshi-etal-2020-state}, we give the count of the languages covered by the classifier in that class, mean F1 score, and mean \ac{fpr} for our model and for that of \citet{costa2022no} (NLLB). 0--5 = least to best resourced.}
    \label{tab:categories}
\end{table}

\subsection{Case study: Chinese languages}
\begin{figure}[tbp]
\small
    \centering
    \includegraphics[width=0.5\textwidth]{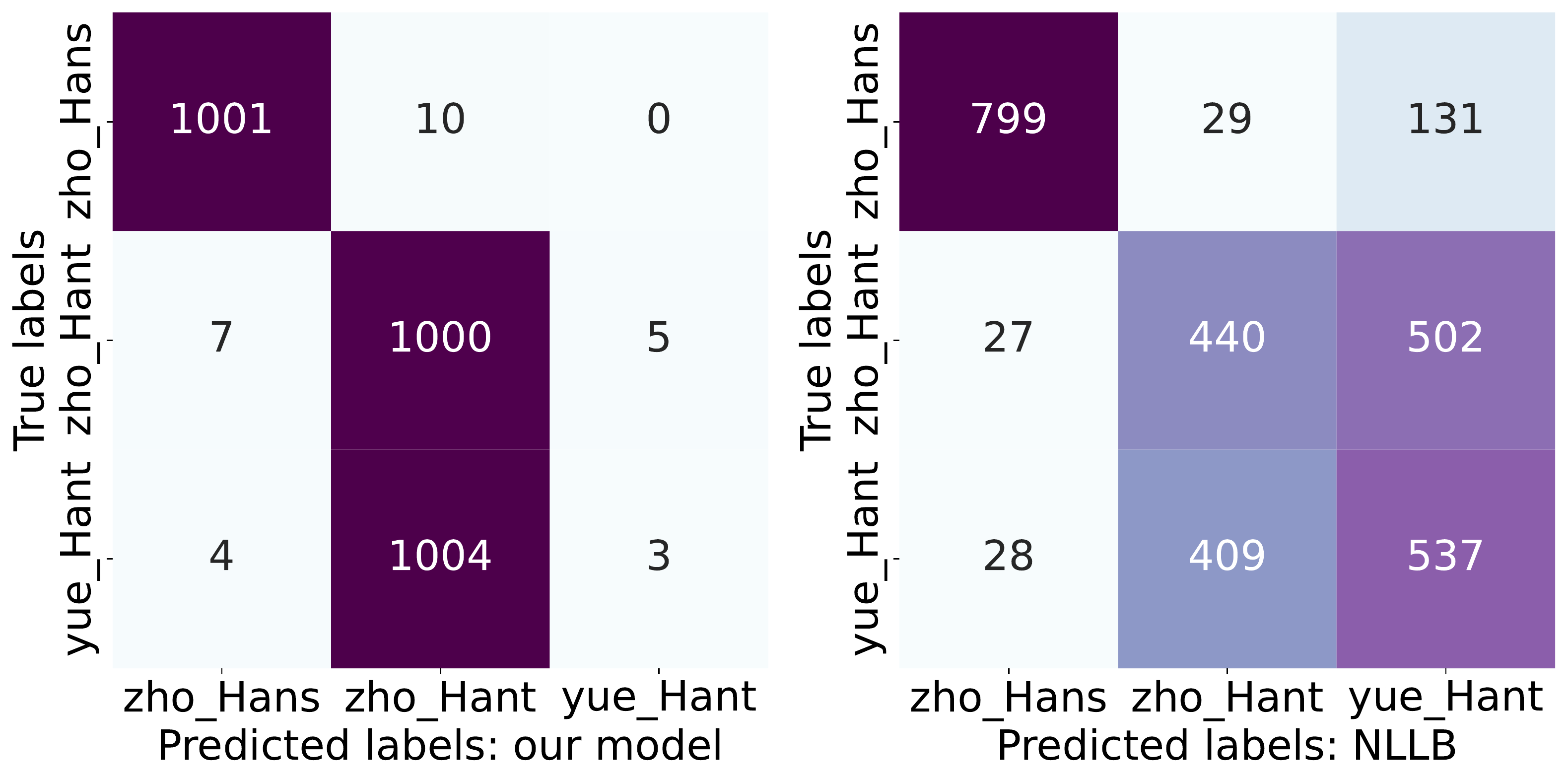}
    \caption{Confusion matrices for our model (L) and NLLB (R), showing the confusion in classification by each model on the FLORES-200 test set between Chinese (Simplified) (\textit{zho\_Hans}), Chinese (Traditional) (\textit{zho\_Hant}), and Yue Chinese (\textit{yue\_Hant}) classes.}
    \label{fig:chinese_cm}
\end{figure}

Despite our model outperforming NLLB overall, NLLB achieved a noticeably higher F1 score on Yue Chinese (0.488 vs. 0.006). \Cref{fig:chinese_cm} shows the confusion matrices for our model and NLLB between the three Chinese languages. Our model performs well on Simplified and Traditional Chinese, but almost never predicts Yue Chinese, instead classifying it as Chinese (Traditional). The NLLB model is also unable to distinguish between Yue and Chinese (Traditional), but mixes the two classes instead.  

We asked four native speakers to inspect our training data and the FLORES-200 test set. They noted that there was a mismatch in domain for Yue Chinese, as much of our training data was written colloquial Yue Chinese whereas the test set consisted of formal writing. Furthermore, they were unable to distinguish with high confidence between Yue and Chinese (Traditional) as the two languages are very similar when written formally. This is an example of a wider problem with \ac{lid}: the language covered by a particular label may vary widely, making single-label classification difficult. 

\section{Conclusion}
We present an open dataset covering 201 languages, which we curate and audit manually to ensure high confidence in its data and language labels. We demonstrate the quality of our dataset by using it to train a high-performing and scalable \ac{lid} model. Finally, we provide detailed analysis into its performance by class. We make both our model and our dataset available to the research community.

\section*{Limitations}
Our dataset and model only covers 201 languages: the ones we were able to test with the FLORES-200 Evaluation Benchmark. In addition, because our test set consists of sentences from a single domain (wiki articles), performance on this test set may not reflect how well our classifier works in other domains. Future work could create a \ac{lid} test set representative of web data where these classifiers are often applied. Finally, most of the data was not audited by native speakers as would be ideal. Future versions of this dataset should have more languages verified by native speakers, with a focus on the least resourced languages.

\section*{Ethics Statement}
Our work aims to broaden \ac{nlp} coverage by allowing practitioners to identify relevant data in more languages. However, we note that \ac{lid} is inherently a normative activity that risks excluding minority dialects, scripts, or entire microlanguages from a macrolanguage. Choosing which languages to cover may reinforce power imbalances, as only some groups gain access to \ac{nlp} technologies.

In addition, errors in \ac{lid} can have a significant impact on downstream performance, particularly (as is often the case) when a system is used as a `black box'. The performance of our classifier is not equal across languages which could lead to worse downstream performance for particular groups. We mitigate this by providing metrics by class.

\section*{Acknowledgements} 
This work was supported in part by the UKRI Centre for Doctoral Training in Natural Language Processing, funded by the UKRI (grant EP/S022481/1) and the University of Edinburgh, School of Informatics and School of Philosophy, Psychology \& Language Sciences.

The experiments in this paper were performed using resources provided by the Cambridge Service for Data Driven Discovery (CSD3) operated by the University of Cambridge Research Computing Service (www.csd3.cam.ac.uk), provided by Dell EMC and Intel using Tier-2 funding from the Engineering and Physical Sciences Research Council (capital grant EP/P020259/1), and DiRAC funding from the Science and Technology Facilities Council (www.dirac.ac.uk). 

Special thanks to Pinzhen Chen, Steven Chien, Bryan Li, Lushi Chen and Victoria Lee for their help with Chinese languages.

\bibliography{anthology,custom}
\bibliographystyle{acl_natbib}

\clearpage
\appendix

\section{Data sources}
\label{sec:sources}

We use the following data sources to build our open dataset. We chose sources as those which were likely to have trustworthy language labels and which did not rely on other \ac{lid} systems for labelling.
\begin{itemize}
    \item Arabic Dialects Dataset \citep{el2018arabic}
    \item Bhojpuri Language Technological Resources Project (BLTR) \citep{ojha2019english}
    \item Global Voices \citep{tiedemann-2012-parallel}
    \item Guaran{\'i} Parallel Set \citep{gongora-etal-2022-use}
    \item The Hong Kong Cantonese corpus (HKCanCor) \citep{luke2015hong}
    \item Integrated dataset for Arabic Dialect Identification (IADD) \citep{zahir2022iadd,alsarsour-etal-2018-dart,abu-kwaik-etal-2018-shami,medhaffar-etal-2017-sentiment,meftouh-etal-2015-machine,zaidan-callison-burch-2011-arabic}
    \item Leipzig Corpora Collection \citep{goldhahn-etal-2012-building}
    \item LTI LangID Corpus \citep{brown2012finding}
    \item MADAR 2019 Shared Task on Arabic Fine-grained Dialect Identification \citep{bouamor-etal-2019-madar}
    \item EM corpus \citep{huidrom-etal-2021-em}
    \item MIZAN \citep{kashefi2018mizan}
    \item MT-560 \citep{gowda-etal-2021-many,tiedemann-2012-parallel,post-etal-2012-constructing,ziemski-etal-2016-united,rozis-skadins-2017-tilde,kunchukuttan-etal-2018-iit,agic-vulic-2019-jw300,espla-etal-2019-paracrawl,qi-etal-2018-pre,zhang-etal-2020-improving,bojar-etal-2013-findings,bojar-etal-2014-findings,bojar-etal-2015-findings,bojar-etal-2016-findings,bojar-etal-2017-findings,bojar-etal-2018-findings,barrault-etal-2019-findings,barrault-etal-2020-findings}
    \item NLLB Seed \citep{costa2022no}
    \item SETIMES news corpus \citep{tiedemann-2012-parallel}
    \item Tatoeba collection \citep{tiedemann-2012-parallel}
    \item Tehran English-Persian Parallel (TEP) Corpus \citep{pilevar2011tep}
    \item Turkish Interlingua (TIL) corpus \citep{mirzakhalov-etal-2021-large}
    \item WiLI benchmark dataset \citep{thoma2018wili}
    \item XL-Sum summarisation dataset \citep{hasan-etal-2021-xl}
\end{itemize}

\section{\ac{lid} model hyperparameters}
\label{sec:hyperparams}

\begin{itemize}
    \item Loss: softmax
    \item Epochs: 2
    \item Learning rate: 0.8
    \item Embedding dimension: 256
    \item Minimum number of word occurences: 1000
    \item Character n-grams: 2--5
    \item Word n-grams: 1
    \item Bucket size: 1,000,000
    \item Threads: 68
\end{itemize}

All other hyperparameters are set to \textit{fasttext} defaults.

\clearpage
\onecolumn

\section{Performance of our \ac{lid} model by language}
\label{sec:by_lang_perf}
\small 
\begin{longtable}{llrrrrr}
    & & & \multicolumn{2}{c}{\textbf{Our model}} & \multicolumn{2}{c}{\textbf{NLLB}} \\
    \cmidrule(lr){4-5} \cmidrule(lr){6-7}
    \textbf{Language code} & \textbf{Language} & \textbf{Training data} & \textbf{F1 score} $ \uparrow $ & \textbf{FPR} $ \downarrow $  & \textbf{F1 score} $ \uparrow $ & \textbf{FPR} $ \downarrow $ \\
    \midrule
    \endhead
    \\
    \caption{For each language covered by our model, we give the number of lines of deduplicated training data in our dataset,  as well as the class F1 score and class false positive rate (FPR) for our model and for the model described in \citet{costa2022no} (NLLB).}\\
    \endfoot
    ace\_Arab & Acehnese & 6191  & 0.9679 & 0.0079 	&	0.9704	&	0.0074	\\
    ace\_Latn & Acehnese & 18032  & 0.9980 & 0.0005 	&	0.9936	&	0.0035	\\
    acm\_Arab & Mesopotamian Arabic & 4862  & 0.0328 & 0.0040 	&	-	&	-	\\
    acq\_Arab & Ta'izzi-Adeni Arabic & 1598  & 0.0020 & 0.0000 	&	-	&	-	\\
    aeb\_Arab & Tunisian Arabic & 18758  & 0.3398 & 0.0479 	&	-	&	-	\\
    afr\_Latn & Afrikaans & 1045638  & 0.9995 & 0.0000 	&	0.9985	&	0.0010	\\
    ajp\_Arab & South Levantine Arabic & 28190  & 0.1906 & 0.0158 	&	-	&	-	\\
    als\_Latn & Tosk Albanian & 506379  & 1.0000 & 0.0000 	&	0.9980	&	0.0020	\\
    amh\_Ethi & Amharic & 606866  & 0.9995 & 0.0005 	&	0.9990	&	0.0010	\\
    apc\_Arab & North Levantine Arabic & 67952  & 0.2334 & 0.0983 	&	-	&	-	\\
    arb\_Arab & Modern Standard Arabic & 7000000  & 0.3077 & 1.1280 	&	0.1903	&	4.2579	\\
    ars\_Arab & Najdi Arabic & 23194  & 0.0184 & 0.1374 	&	-	&	-	\\
    ary\_Arab & Moroccan Arabic & 25411  & 0.4894 & 0.7643 	&	-	&	-	\\
    arz\_Arab & Egyptian Arabic & 52327  & 0.4235 & 1.0875 	&	-	&	-	\\
    asm\_Beng & Assamese & 161726  & 1.0000 & 0.0000 	&	1.0000	&	0.0000	\\
    ast\_Latn & Asturian & 35815  & 0.9901 & 0.0045 	&	0.9902	&	0.0069	\\
    awa\_Deva & Awadhi & 4957  & 0.6770 & 0.0040 	&	0.9611	&	0.0084	\\
    ayr\_Latn & Central Aymara & 142628  & 1.0000 & 0.0000 	&	0.9980	&	0.0005	\\
    azb\_Arab & South Azerbaijani & 532  & 0.7514 & 0.0000 	&	0.8805	&	0.0069	\\
    azj\_Latn & North Azerbaijani & 462672  & 0.9990 & 0.0005 	&	0.9970	&	0.0030	\\
    bak\_Cyrl & Bashkir & 65942  & 1.0000 & 0.0000 	&	0.9990	&	0.0005	\\
    bam\_Latn & Bambara & 9538  & 0.6107 & 0.4926 	&	0.6194	&	0.4826	\\
    ban\_Latn & Balinese & 15404  & 0.9789 & 0.0015 	&	0.9712	&	0.0030	\\
    bel\_Cyrl & Belarusian & 84846  & 1.0000 & 0.0000 	&	1.0000	&	0.0000	\\
    bem\_Latn & Bemba & 383559  & 0.9796 & 0.0193 	&	0.9739	&	0.0252	\\
    ben\_Beng & Bengali & 490226  & 0.9925 & 0.0000 	&	0.9995	&	0.0005	\\
    bho\_Deva & Bhojpuri & 69367  & 0.8921 & 0.1136 	&	0.9335	&	0.0153	\\
    bjn\_Arab & Banjar & 6192  & 0.9604 & 0.0257 	&	0.9524	&	0.0163	\\
    bjn\_Latn & Banjar & 21475  & 0.9857 & 0.0064 	&	0.8336	&	0.1721	\\
    bod\_Tibt & Standard Tibetan & 2514  & 0.8045 & 0.0000 	&	0.9637	&	0.0366	\\
    bos\_Latn & Bosnian & 330473  & 0.6928 & 0.0939 	&	0.5954	&	0.0584	\\
    bug\_Latn & Buginese & 7527  & 0.9970 & 0.0005 	&	0.9765	&	0.0054	\\
    bul\_Cyrl & Bulgarian & 610545  & 1.0000 & 0.0000 	&	0.9995	&	0.0000	\\
    cat\_Latn & Catalan & 115963  & 1.0000 & 0.0000 	&	0.9873	&	0.0129	\\
    ceb\_Latn & Cebuano & 1002342  & 0.9995 & 0.0005 	&	0.9995	&	0.0000	\\
    ces\_Latn & Czech & 424828  & 0.9975 & 0.0015 	&	0.9990	&	0.0010	\\
    cjk\_Latn & Chokwe & 36244  & 0.9023 & 0.0025 	&	0.8688	&	0.0089	\\
    ckb\_Arab & Central Kurdish & 17792  & 1.0000 & 0.0000 	&	1.0000	&	0.0000	\\
    crh\_Latn & Crimean Tatar & 19148  & 0.9920 & 0.0005 	&	0.9829	&	0.0000	\\
    cym\_Latn & Welsh & 98719  & 1.0000 & 0.0000 	&	1.0000	&	0.0000	\\
    dan\_Latn & Danish & 2789406  & 0.9881 & 0.0035 	&	0.9946	&	0.0020	\\
    deu\_Latn & German & 653914  & 1.0000 & 0.0000 	&	0.9907	&	0.0094	\\
    dik\_Latn & Southwestern Dinka & 25911  & 0.9995 & 0.0000 	&	0.9925	&	0.0000	\\
    dyu\_Latn & Dyula & 17351  & 0.0421 & 0.0282 	&	0.0480	&	0.0228	\\
    dzo\_Tibt & Dzongkha & 6899  & 0.8585 & 0.1635 	&	0.9679	&	0.0005	\\
    ell\_Grek & Greek & 3312774  & 1.0000 & 0.0000 	&	1.0000	&	0.0000	\\
    eng\_Latn & English & 7544560  & 0.9941 & 0.0049 	&	0.9792	&	0.0213	\\
    epo\_Latn & Esperanto & 339280  & 1.0000 & 0.0000 	&	0.9970	&	0.0030	\\
    est\_Latn & Estonian & 3331470  & 0.9990 & 0.0005 	&	0.9985	&	0.0015	\\
    eus\_Latn & Basque & 622029  & 0.9990 & 0.0005 	&	0.9985	&	0.0015	\\
    ewe\_Latn & Ewe & 585267  & 0.9980 & 0.0020 	&	0.9970	&	0.0030	\\
    fao\_Latn & Faroese & 40022  & 1.0000 & 0.0000 	&	0.5052	&	0.0000	\\
    fij\_Latn & Fijian & 360981  & 0.9985 & 0.0005 	&	1.0000	&	0.0000	\\
    fin\_Latn & Finnish & 2613970  & 0.9995 & 0.0005 	&	0.9995	&	0.0005	\\
    fon\_Latn & Fon & 31875  & 0.9980 & 0.0000 	&	0.9970	&	0.0000	\\
    fra\_Latn & French & 586938  & 0.9950 & 0.0000 	&	0.9961	&	0.0035	\\
    fur\_Latn & Friulian & 55622  & 0.9985 & 0.0015 	&	0.9980	&	0.0000	\\
    fuv\_Latn & Nigerian Fulfulde & 14419  & 0.9865 & 0.0005 	&	0.9810	&	0.0040	\\
    gaz\_Latn & West Central Oromo & 335769  & 0.9990 & 0.0010 	&	0.9995	&	0.0005	\\
    gla\_Latn & Scottish Gaelic & 52665  & 0.9975 & 0.0025 	&	0.9985	&	0.0010	\\
    gle\_Latn & Irish & 211460  & 1.0000 & 0.0000 	&	0.9980	&	0.0020	\\
    glg\_Latn & Galician & 42017  & 0.9970 & 0.0025 	&	0.9931	&	0.0049	\\
    grn\_Latn & Guarani & 57458  & 0.9975 & 0.0025 	&	0.9965	&	0.0015	\\
    guj\_Gujr & Gujarati & 836618  & 1.0000 & 0.0000 	&	1.0000	&	0.0000	\\
    hat\_Latn & Haitian Creole & 299853  & 0.9970 & 0.0030 	&	0.9985	&	0.0005	\\
    hau\_Latn & Hausa & 347741  & 0.9893 & 0.0109 	&	0.9970	&	0.0025	\\
    heb\_Hebr & Hebrew & 944918  & 0.9990 & 0.0010 	&	1.0000	&	0.0000	\\
    hin\_Deva & Hindi & 1089471  & 0.8477 & 0.1749 	&	0.8722	&	0.1454	\\
    hne\_Deva & Chhattisgarhi & 52819  & 0.9362 & 0.0311 	&	0.9300	&	0.0134	\\
    hrv\_Latn & Croatian & 832967  & 0.7441 & 0.1863 	&	0.7335	&	0.2645	\\
    hun\_Latn & Hungarian & 2870535  & 1.0000 & 0.0000 	&	0.9926	&	0.0074	\\
    hye\_Armn & Armenian & 368832  & 1.0000 & 0.0000 	&	1.0000	&	0.0000	\\
    ibo\_Latn & Igbo & 491594  & 0.9995 & 0.0005 	&	0.9995	&	0.0005	\\
    ilo\_Latn & Ilocano & 976648  & 0.9990 & 0.0010 	&	0.9985	&	0.0015	\\
    ind\_Latn & Indonesian & 1694230  & 0.9279 & 0.0435 	&	0.8198	&	0.2087	\\
    isl\_Latn & Icelandic & 43554  & 1.0000 & 0.0000 	&	0.7621	&	0.3125	\\
    ita\_Latn & Italian & 479663  & 0.9940 & 0.0000 	&	0.9721	&	0.0282	\\
    jav\_Latn & Javanese & 65595  & 0.9917 & 0.0079 	&	0.9767	&	0.0218	\\
    jpn\_Jpan & Japanese & 876783  & 1.0000 & 0.0000 	&	0.9808	&	0.0104	\\
    kab\_Latn & Kabyle & 52634  & 0.8551 & 0.1695 	&	0.8579	&	0.1652	\\
    kac\_Latn & Jingpho & 11365  & 1.0000 & 0.0000 	&	1.0000	&	0.0000	\\
    kam\_Latn & Kamba & 52674  & 0.9001 & 0.0005 	&	0.7581	&	0.0010	\\
    kan\_Knda & Kannada & 357780  & 1.0000 & 0.0000 	&	1.0000	&	0.0000	\\
    kas\_Arab & Kashmiri & 6203  & 0.9839 & 0.0000 	&	0.9710	&	0.0000	\\
    kas\_Deva & Kashmiri & 6694  & 0.9860 & 0.0010 	&	0.9840	&	0.0005	\\
    kat\_Geor & Georgian & 417604  & 1.0000 & 0.0000 	&	1.0000	&	0.0000	\\
    kaz\_Cyrl & Kazakh & 51577  & 0.9995 & 0.0000 	&	0.9995	&	0.0000	\\
    kbp\_Latn & Kabiye & 53275  & 1.0000 & 0.0000 	&	1.0000	&	0.0000	\\
    kea\_Latn & Kabuverdianu & 5665  & 0.9652 & 0.0000 	&	0.9610	&	0.0000	\\
    khk\_Cyrl & Halh Mongolian & 168540  & 1.0000 & 0.0000 	&	1.0000	&	0.0000	\\
    khm\_Khmr & Khmer & 60513  & 0.9995 & 0.0000 	&	0.9990	&	0.0000	\\
    kik\_Latn & Kikuyu & 96402  & 0.9628 & 0.0376 	&	0.9636	&	0.0341	\\
    kin\_Latn & Kinyarwanda & 447057  & 0.8872 & 0.0069 	&	0.9788	&	0.0119	\\
    kir\_Cyrl & Kyrgyz & 372399  & 1.0000 & 0.0000 	&	1.0000	&	0.0000	\\
    kmb\_Latn & Kimbundu & 92635  & 0.9394 & 0.0534 	&	0.9361	&	0.0514	\\
    kmr\_Latn & Northern Kurdish & 15490  & 0.9985 & 0.0010 	&	0.9956	&	0.0045	\\
    knc\_Arab & Central Kanuri & 6196  & 0.7017 & 0.0000 	&	0.7026	&	0.0000	\\
    knc\_Latn & Central Kanuri & 6256  & 0.9990 & 0.0005 	&	0.9965	&	0.0015	\\
    kon\_Latn & Kikongo & 209801  & 0.9946 & 0.0045 	&	0.9936	&	0.0049	\\
    kor\_Hang & Korean & 1772136  & 1.0000 & 0.0000 	&	0.9961	&	0.0040	\\
    lao\_Laoo & Lao & 23529  & 1.0000 & 0.0000 	&	0.9995	&	0.0000	\\
    lij\_Latn & Ligurian & 28641  & 0.9980 & 0.0015 	&	0.9774	&	0.0025	\\
    lim\_Latn & Limburgish & 48151  & 0.9965 & 0.0015 	&	0.9870	&	0.0010	\\
    lin\_Latn & Lingala & 546344  & 0.9990 & 0.0010 	&	0.9956	&	0.0030	\\
    lit\_Latn & Lithuanian & 2663659  & 0.9985 & 0.0010 	&	0.9990	&	0.0010	\\
    lmo\_Latn & Lombard & 35402  & 0.9975 & 0.0020 	&	0.9696	&	0.0109	\\
    ltg\_Latn & Latgalian & 15585  & 0.9985 & 0.0000 	&	0.9920	&	0.0000	\\
    ltz\_Latn & Luxembourgish & 37674  & 0.9995 & 0.0000 	&	0.9995	&	0.0000	\\
    lua\_Latn & Luba-Kasai & 292972  & 0.9960 & 0.0005 	&	0.9936	&	0.0035	\\
    lug\_Latn & Ganda & 251105  & 0.9941 & 0.0045 	&	0.9921	&	0.0069	\\
    luo\_Latn & Luo & 138159  & 0.9985 & 0.0015 	&	0.9975	&	0.0005	\\
    lus\_Latn & Mizo & 195262  & 0.9985 & 0.0000 	&	0.9945	&	0.0005	\\
    lvs\_Latn & Standard Latvian & 2872096  & 0.9990 & 0.0005 	&	0.9936	&	0.0064	\\
    mag\_Deva & Magahi & 6208  & 0.9620 & 0.0133 	&	0.9311	&	0.0213	\\
    mai\_Deva & Maithili & 15385  & 0.9880 & 0.0010 	&	0.9871	&	0.0040	\\
    mal\_Mlym & Malayalam & 379786  & 1.0000 & 0.0000 	&	1.0000	&	0.0000	\\
    mar\_Deva & Marathi & 1017951  & 0.9990 & 0.0010 	&	0.9951	&	0.0049	\\
    min\_Latn & Minangkabau & 31469  & 0.9931 & 0.0030 	&	0.5143	&	0.0010	\\
    mkd\_Cyrl & Macedonian & 561725  & 0.9995 & 0.0005 	&	1.0000	&	0.0000	\\
    mlt\_Latn & Maltese & 2219213  & 0.9985 & 0.0015 	&	0.9995	&	0.0005	\\
    mni\_Beng & Meitei & 47146  & 0.9941 & 0.0059 	&	0.9995	&	0.0000	\\
    mos\_Latn & Mossi & 197187  & 0.9814 & 0.0005 	&	0.9684	&	0.0000	\\
    mri\_Latn & Maori & 48792  & 0.9995 & 0.0005 	&	0.9985	&	0.0005	\\
    mya\_Mymr & Burmese & 452194  & 1.0000 & 0.0000 	&	1.0000	&	0.0000	\\
    nld\_Latn & Dutch & 2929602  & 0.9970 & 0.0015 	&	0.9830	&	0.0173	\\
    nno\_Latn & Norwegian Nynorsk & 101140  & 0.9828 & 0.0104 	&	0.9697	&	0.0208	\\
    nob\_Latn & Norwegian Bokmal & 1783598  & 0.9719 & 0.0148 	&	0.9829	&	0.0139	\\
    npi\_Deva & Nepali & 60345  & 0.9980 & 0.0020 	&	0.9980	&	0.0020	\\
    nso\_Latn & Northern Sotho & 560068  & 0.9868 & 0.0119 	&	0.9839	&	0.0134	\\
    nus\_Latn & Nuer & 6295  & 0.9995 & 0.0000 	&	0.9980	&	0.0015	\\
    nya\_Latn & Nyanja & 789078  & 0.9966 & 0.0035 	&	0.9460	&	0.0163	\\
    oci\_Latn & Occitan & 32683  & 0.9941 & 0.0054 	&	0.9835	&	0.0163	\\
    ory\_Orya & Odia & 92355  & 1.0000 & 0.0000 	&	1.0000	&	0.0000	\\
    pag\_Latn & Pangasinan & 294618  & 0.9990 & 0.0005 	&	0.9970	&	0.0010	\\
    pan\_Guru & Eastern Panjabi & 357487  & 1.0000 & 0.0000 	&	1.0000	&	0.0000	\\
    pap\_Latn & Papiamento & 403991  & 0.9768 & 0.0232 	&	0.9839	&	0.0158	\\
    pbt\_Arab & Southern Pasto & 63256  & 0.9980 & 0.0015 	&	0.9970	&	0.0010	\\
    pes\_Arab & Western Persian & 1758215  & 0.5570 & 0.5356 	&	0.6385	&	0.4381	\\
    plt\_Latn & Plateau Malgasy & 47284  & 1.0000 & 0.0000 	&	1.0000	&	0.0000	\\
    pol\_Latn & Polish & 3403455  & 0.9956 & 0.0045 	&	0.9849	&	0.0153	\\
    por\_Latn & Portuguese & 3800360  & 0.9941 & 0.0040 	&	0.9854	&	0.0143	\\
    prs\_Arab & Dari & 6662  & 0.5144 & 0.1122 	&	0.4589	&	0.0608	\\
    quy\_Latn & Ayacucho Quechua & 154448  & 1.0000 & 0.0000 	&	1.0000	&	0.0000	\\
    ron\_Latn & Romanian & 443200  & 0.9985 & 0.0015 	&	0.9985	&	0.0015	\\
    run\_Latn & Rundi & 459617  & 0.9044 & 0.0973 	&	0.9782	&	0.0104	\\
    rus\_Cyrl & Russian & 7000000  & 0.9990 & 0.0005 	&	0.9990	&	0.0010	\\
    sag\_Latn & Sango & 255491  & 0.9990 & 0.0000 	&	0.9970	&	0.0005	\\
    san\_Deva & Sanskrit & 39988  & 0.9900 & 0.0000 	&	0.9885	&	0.0010	\\
    sat\_Olck & Santali & 8875  & 1.0000 & 0.0000 	&	1.0000	&	0.0000	\\
    scn\_Latn & Sicilian & 40023  & 0.9956 & 0.0035 	&	0.9936	&	0.0054	\\
    shn\_Mymr & Shan & 21051  & 1.0000 & 0.0000 	&	0.9985	&	0.0000	\\
    sin\_Sinh & Sinhala & 361636  & 1.0000 & 0.0000 	&	1.0000	&	0.0000	\\
    slk\_Latn & Slovak & 3153492  & 0.9970 & 0.0010 	&	0.9995	&	0.0005	\\
    slv\_Latn & Slovenian & 3023266  & 0.9966 & 0.0030 	&	0.9985	&	0.0015	\\
    smo\_Latn & Samoan & 367828  & 0.9985 & 0.0010 	&	0.9985	&	0.0010	\\
    sna\_Latn & Shona & 764419  & 0.9941 & 0.0059 	&	0.9941	&	0.0059	\\
    snd\_Arab & Sindhi & 26107  & 0.9990 & 0.0000 	&	0.9980	&	0.0020	\\
    som\_Latn & Somali & 217413  & 0.9995 & 0.0005 	&	1.0000	&	0.0000	\\
    sot\_Latn & Southern Sotho & 2030  & 0.9567 & 0.0000 	&	0.7552	&	0.0000	\\
    spa\_Latn & Spanish & 677548  & 0.9921 & 0.0049 	&	0.9922	&	0.0074	\\
    srd\_Latn & Sardinian & 47480  & 0.9961 & 0.0030 	&	0.9773	&	0.0000	\\
    srp\_Cyrl & Serbian & 310259  & 0.9995 & 0.0000 	&	1.0000	&	0.0000	\\
    ssw\_Latn & Swati & 114900  & 0.9911 & 0.0020 	&	0.9916	&	0.0015	\\
    sun\_Latn & Sundanese & 47458  & 0.9926 & 0.0035 	&	0.9599	&	0.0252	\\
    swe\_Latn & Swedish & 2747052  & 1.0000 & 0.0000 	&	0.9990	&	0.0005	\\
    swh\_Latn & Swahili & 228559  & 0.9284 & 0.0771 	&	0.8815	&	0.1345	\\
    szl\_Latn & Silesian & 34065  & 0.9960 & 0.0000 	&	0.9875	&	0.0015	\\
    tam\_Taml & Tamil & 552180  & 1.0000 & 0.0000 	&	1.0000	&	0.0000	\\
    taq\_Latn & Tamasheq & 10266  & 0.7907 & 0.0010 	&	0.7916	&	0.0000	\\
    taq\_Tfng & Tamasheq & 6203  & 0.9505 & 0.0084 	&	0.8513	&	0.0000	\\
    tat\_Cyrl & Tatar & 257828  & 1.0000 & 0.0000 	&	0.9995	&	0.0000	\\
    tel\_Telu & Telugu & 276504  & 0.9990 & 0.0000 	&	1.0000	&	0.0000	\\
    tgk\_Cyrl & Tajik & 135652  & 1.0000 & 0.0000 	&	1.0000	&	0.0000	\\
    tgl\_Latn & Tagalog & 1189616  & 1.0000 & 0.0000 	&	0.9970	&	0.0025	\\
    tha\_Thai & Thai & 734727  & 1.0000 & 0.0000 	&	1.0000	&	0.0000	\\
    tir\_Ethi & Tigrinya & 333639  & 0.9995 & 0.0000 	&	0.9995	&	0.0000	\\
    tpi\_Latn & Tok Pisin & 471651  & 1.0000 & 0.0000 	&	0.9980	&	0.0000	\\
    tsn\_Latn & Tswana & 784851  & 0.9693 & 0.0311 	&	0.8424	&	0.1859	\\
    tso\_Latn & Tsonga & 756533  & 0.9961 & 0.0035 	&	0.9907	&	0.0089	\\
    tuk\_Latn & Turkmen & 160757  & 1.0000 & 0.0000 	&	1.0000	&	0.0000	\\
    tum\_Latn & Tumbuka & 237138  & 0.9956 & 0.0035 	&	0.9816	&	0.0183	\\
    tur\_Latn & Turkish & 823575  & 0.9936 & 0.0064 	&	0.9840	&	0.0163	\\
    twi\_Latn & Twi & 545217  & 0.9990 & 0.0000 	&	0.9420	&	0.0005	\\
    tzm\_Tfng & Central Atlas Tamazight & 8142  & 0.9535 & 0.0395 	&	0.8854	&	0.1296	\\
    uig\_Arab & Uyghur & 57231  & 1.0000 & 0.0000 	&	0.9995	&	0.0005	\\
    ukr\_Cyrl & Ukrainian & 1140463  & 0.9995 & 0.0005 	&	1.0000	&	0.0000	\\
    umb\_Latn & Umbundu & 220396  & 0.9776 & 0.0079 	&	0.9687	&	0.0208	\\
    urd\_Arab & Urdu & 412736  & 0.9849 & 0.0153 	&	0.9735	&	0.0272	\\
    uzn\_Latn & Northern Uzbek & 1519230  & 0.9990 & 0.0010 	&	0.9995	&	0.0005	\\
    vec\_Latn & Venetian & 43478  & 0.9961 & 0.0020 	&	0.9916	&	0.0035	\\
    vie\_Latn & Vietnamese & 881145  & 0.9995 & 0.0005 	&	0.9873	&	0.0129	\\
    war\_Latn & Waray & 282772  & 1.0000 & 0.0000 	&	0.9990	&	0.0010	\\
    wol\_Latn & Wolof & 28784  & 0.9970 & 0.0020 	&	0.9950	&	0.0010	\\
    xho\_Latn & Xhosa & 921590  & 0.9858 & 0.0119 	&	0.9779	&	0.0148	\\
    ydd\_Hebr & Eastern Yiddish & 911  & 0.9990 & 0.0000 	&	1.0000	&	0.0000	\\
    yor\_Latn & Yoruba & 531904  & 0.9990 & 0.0010 	&	0.9956	&	0.0030	\\
    yue\_Hant & Yue Chinese & 63254  & 0.0059 & 0.0025 	&	0.4877	&	0.3229	\\
    zho\_Hans & Chinese (Simplified) & 1046823  & 0.9891 & 0.0054 	&	0.8559	&	0.0277	\\
    zho\_Hant & Chinese (Traditional) & 2018541  & 0.6605 & 0.5020 	&	0.4651	&	0.2176	\\
    zsm\_Latn & Standard Malay & 404380  & 0.9495 & 0.0346 	&	0.9351	&	0.0307	\\
    zul\_Latn & Zulu & 951688  & 0.9828 & 0.0104 	&	0.9696	&	0.0267	\\
   
\end{longtable}

\end{document}